%% file: main.tex
\documentclass[letterpaper, 10 pt, conference]{ieeeconf}
\IEEEoverridecommandlockouts
\usepackage{graphicx}
\usepackage{caption}
\graphicspath{
    {graphics/}
}

\usepackage{amsmath,amssymb,enumerate}

\usepackage{amsthm}
\usepackage{setspace}
\usepackage{booktabs}
\usepackage{xcolor}
\usepackage{hyperref}
\hypersetup{
  hypertexnames= false,
  breaklinks   = true,
  plainpages   = true,
  hidelinks    = true,
  colorlinks   = true, 
  urlcolor     = red, 
  linkcolor    = black, 
  citecolor    = black 
}
\usepackage{blindtext}
\usepackage{gensymb}
\usepackage{xparse}
\usepackage{lipsum}
\usepackage{soul}
\usepackage{mathrsfs}
\usepackage[mathscr]{euscript}
\usepackage{times}
\usepackage{cite} 
\usepackage{multicol}
\usepackage[caption=false,font=footnotesize]{subfig}
\usepackage{amsfonts}
\usepackage{fixltx2e}
\usepackage{cleveref}
\usepackage[utf8]{inputenc}
\usepackage[T1]{fontenc}
\usepackage{textcomp}
\usepackage{balance}
\usepackage{siunitx}
\usepackage{comment}
\usepackage[normalem]{ulem}

\mathchardef\mhyphen="2D 

\newcommand{\RTSimulink}{\texttt{Real-Time Simulink}~}
\newcommand{\SimMech}{\texttt{SimMechanics}~}

\newcommand{\remark}[1]{\noindent \textbf{Remark:} \emph{#1}}
\newcommand{\squeezeup}{\vspace{-3mm}}

\title{\LARGE \bf
Feedback Control of a Cassie Bipedal Robot: \\
Walking, Standing, and Riding a Segway
}

\author{Yukai Gong, Ross Hartley, Xingye Da, Ayonga Hereid, Omar Harib, Jiunn-Kai Huang, and Jessy Grizzle
\thanks{The authors are with the College of Engineering and the Robotics Institute, University of Michigan, Ann Arbor, MI 48109 USA {\tt\small \{ykgong,rosshart,xda,ayonga,oharib,bjhuang,
 grizzle\}}@umich.edu }
}

\begin{document}
\maketitle
\begin{abstract}
The Cassie bipedal robot designed by Agility Robotics is providing academics a common platform for sharing and comparing algorithms for locomotion, perception, and navigation. This paper focuses on feedback control for standing and walking using the methods of virtual constraints and gait libraries. The designed controller was implemented six weeks after the robot arrived at the University of Michigan and allowed it to stand in place as well as walk over sidewalks, grass, snow, sand, and burning brush. The controller for standing also enables the robot to ride a Segway. A model of the Cassie robot has been placed on GitHub and the controller will also be made open source if the paper is accepted.
\end{abstract}

\thispagestyle{empty}
\pagestyle{empty}
\pagestyle{plain}

\input{Sections/Introduction}

\input{Sections/CassieRobot}

\input{Sections/WalkingController}

\input{Sections/Standing}

\input{Sections/Conclusion}


\section*{ACKNOWLEDGMENT}
\footnotesize{
The authors thank S. Danforth for insightful comments. Agility Robotics designed and built the robot. Funding for this work was provided in part by the Toyota Research Institute (TRI) under award number No.~02281 and in part by NSF Award No.~1525006. All opinions are those of the authors.}

\bibliographystyle{plain}
\bibliography{BibFiles/Bib2018July.bib,BibFiles/bib2.bib}
\end{document}

%% file: Sections/Introduction.tex
\section{Introduction and Related Work}
\label{sec:intro}

\begin{figure}[b!]
\squeezeup
	\centering
	\includegraphics[width=0.9\columnwidth]{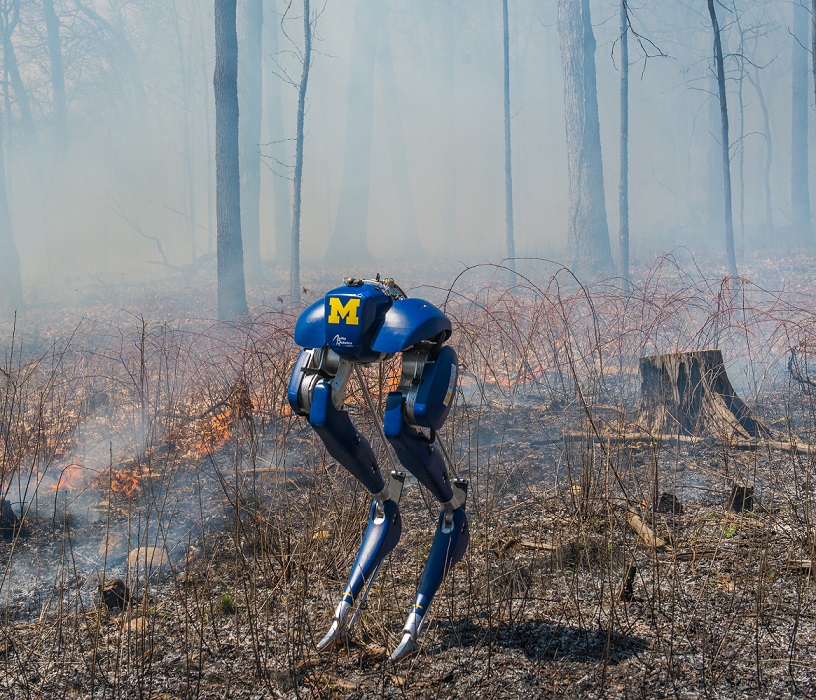}
	\caption{Cassie Blue, one of Michigan's Cassie-series robots designed by Agility Robotics. The robot is shown here participating in a controlled burn. It has 20 degrees of freedom, 10 actuators, joint encoders, and an IMU. The robot's serial number is 001.}
	\label{fig:CassieBlueFire}
\end{figure}

The primary purpose of this paper is to introduce and document an \textit{open-source  controller} for the Cassie bipedal robot shown in Fig.~\ref{fig:CassieBlueFire} and video \cite{CassieACC2019}. Agility Robotics delivered its first batch of Cassie robots to the University of Michigan, Caltech, and Harvard at the end of August, beginning of September, 2017. Since then, each team has been working to develop its own control laws \cite{xiong2018bipedal}. 

\subsection{Contributions}
The Cassie robot comes with a proprietary control law in the form of \texttt{.p MATLAB files} (i.e., code is obfuscated) that run in \RTSimulink. The native controller provides for quiet standing, walking forward, backward, and sideways, and turning. 
The controller presented here provides these same features. 

Among this paper's contributions are (a) the controller's design is documented and (b) its implementation in \RTSimulink will be available on GitHub if the paper is accepted. Additional contributions include the following: (c) The robot is operated in more challenging conditions than those demonstrated by Agility Robotics, such as walking in a controlled burn, walking in snow, and walking in soft sand \cite{CassieACC2019}. (d) The robot is treated as being underactuated in the sense that stance ankle torque is not required to maintain balance when walking. Because of this, the robot's feet sinking into soft sand at a 30$^\degree$ pitch angle does not affect stability of the gait. (e) Finally, the robot is demonstrated riding a Segway, a task relevant to search and rescue operations where a robot could use a tool found at the scene to save energy or effect a faster arrival. 

\subsection{Related Work}
  

The majority of walking gaits in 3D humanoid robots are stabilized by regulating the Zero-Moment Point (ZMP) \cite{hirai1998development,huang1999high,kajita2003biped,dai2016planning} (roughly the center of pressure of the stance foot). Because this method requires having a non-zero support polygon, it is not applicable to robots such as Cassie with line feet \cite{VUBO04}. Simplifying the robot's dynamics to a point mass attached to a rod of fixed length, the so-called (inverted) pendulum models, is another ubiquitous means in the bipedal robotics literature to approach the gait design and stabilization problem \cite{robots_HRP4,PRKODBRECOJONE12, KrEnWiOt12, FaPoAtIj14, ReHuJoPeVaJoAbHu15}. The many challenges associated with the pendulum approach include, exploiting the full capability of the robot, respecting motor and joint limitations, and deciding how to associate the states of the full-order system with the motion of the low-dimensional pendulum. 


An important feature of the pendulum models, nevertheless, is that the Cartesian velocity of their center of mass can be ``stabilized by foot placement''; more precisely, by controlling the angle of the swing leg at ground contact, the robot's speed can be regulated to a desired value \cite{Pratt200,KODBREGOPR12,schepelmannexperimental}. Indeed, the open-loop center of mass velocity of the linear inverted pendulum (LIP) evolves as an integrator in the longitudinal and lateral directions and can therefore be stabilized with a proportional controller.

The controller presented here is based on the \textit{gait library method} introduced in  \cite{da20162d}. It allows the full dynamic model of the robot to be used for the design of periodic walking gaits so that key limits on joint angles, ground reaction forces, motor torque, and motor power can be directly incorporated as constraints into a parameter optimization problem that minimizes energy consumption per distance traveled for a given speed. The individual periodic gaits are realized on the robot via virtual constraints \cite{CA04}, which are in turn ``gain scheduled'' with the robot's current velocity, thereby creating a continuum of gaits. The resulting closed-loop system enjoys the property that its center of mass velocity evolves \textit{approximately} as an integrator in the longitudinal and lateral directions \cite{da20162d}, and hence, similar to the inverted pendulum models, walking speed can be regulated to a desired value via ``foot placement'' as in \cite{rezazadeh2015spring}.

%% file: Sections/CassieRobot.tex
\section{Cassie Robot}
\label{sec:CassieRobot}

The bipedal robot shown in Fig.~\ref{fig:CassieBlueFire}, named Cassie Blue, is designed and built by Agility Robotics. It is all electric and can walk for approximately four hours on a single battery charge. The robot's morphology and name are inspired by the \textit{Cassowary}, a flightless bird native to New Guinea; those without a biology background can liken it to an ostrich. 

\begin{figure}[b]
\vspace{-2mm}
	\centering
	\includegraphics[width=0.7\columnwidth]{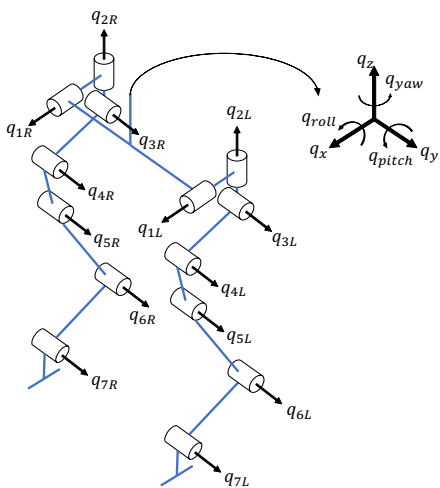}
	\caption{A kinematic model of Cassie. The shin is the link between joints $q_5$ and $q_6$. The tarsus is the link between joints $q_6$ and $q_7$. The last link on the robot is anatomically a toe, but we will call it a foot because it functions as one. The joints $q_1$, $\ldots$, $q_4$ and $q_7$ are actuated and corresponding the motor torques are labeled $u_1$, $\ldots$, $u_4$ and $u_5$ in the obvious way.}
	\label{fig:CanModel}
\squeezeup \squeezeup
\end{figure}

When standing up, Cassie is about one meter tall. The robot's total mass is 31 kg, of which the battery, housed in its torso, accounts for approximately 4 kg.  The floating base model of the robot has 20 DoF. There are seven joints in each leg, five of which are actuated by electric motors and the other two are passive joints realized via a four-bar linkage where one of the links is a leaf spring \cite[Fig. 2]{xiong2018bipedal}. Three of the five actuated joints in each leg correspond to setting the roll, yaw, and pitch of the leg with respect to the torso, one motor sets the angle of the ``knee'', and the last motor regulates the pitch angle of the toe, hereafter called the foot. The robot is fully capable of 3D walking and quiet standing.

\subsection{Coordinate and Actuation Description}
\label{sec:CassieDescription}

A URDF of the robot Cassie is available online \cite{cassiemodel}; in addition, Agility Robotics has released two dynamic models of the robot \cite{cassiesims}. Figure~\ref{fig:CanModel} shows a kinematic model of Cassie with the configuration variables labeled. The variables $(q_x, q_y, q_z) $ are the Cartesian position of the torso, and $q_{yaw}, q_{pitch}, q_{roll}$ are the intrinsic Euler Angles taken in the $ z\mhyphen y\mhyphen x $ order. The coordinates $q_1$, $q_2$, $q_3$ are the hip roll, hip yaw, and hip pitch angles, respectively, $q_4$ is the knee pitch angle, $q_5$ is the shin pitch angle, $q_6$ is the tarsus pitch angle, and $q_7$ is the toe pitch angle. Very roughly speaking, in a human, the tarsus is the part of the foot from the heel to the metatarsus, which in turn is the part of the foot from which the toes radiate. When we walk on two legs, our tarsus is on the ground and we call it the foot. When we crawl on all fours, our tarsus is in the air, like Cassie's. From now on, we'll also call Cassie's toe a ``foot'' because it functions as a foot with respect to most other bipedal robots.

The coordinates are lumped into a vector 
\begin{equation}
\label{eq:qdef}
\begin{aligned}
q:= &[q_x, q_y, q_z, q_{\rm yaw}, q_{\rm pitch}, q_{\rm roll}, \\
&~q_{1L}, q_{2L}, q_{3L}, q_{4L}, q_{5L}, q_{6L}, q_{7L}, \\
&~q_{1R}, q_{2R}, q_{3R}, q_{4R}, q_{5R}, q_{6R}, q_{7R}]^{\mathsf{T}}.
\end{aligned}
\end{equation}
The \textit{body} or \textit{shape} coordinates are
\begin{equation}
\label{eq:qbody}
q_{\rm body}:= [q_{1L}, \ldots, q_{7L}, q_{1R}, \dots, q_{7R}]^{\mathsf{T}},
\end{equation}
the relative angles of the joints connecting the various links of the robot; these are measured via encoders while the Euler parameters in $q$ are estimated by an EKF that comes with the IMU on Cassie, a VectorNav-100.

The joints $(q_1, q_2, q_3, q_4, q_7)$ are actuated. The joints $(q_5, q_6)$ are passive, being formed by the introduction of (rather stiff) springs. When the springs are uncompressed, 
\begin{equation}
\label{eq:SpringRelaxed}
q_5=0,~ \text{and}~q_6=-q_4+ 13^\degree.
\end{equation}


\squeezeup
\subsection{Floating-base Model}
\label{sec:CassieModelFB}
The floating-base dynamic model is expressed in the form of Lagrange as 
\begin{equation}
\label{eq:FloatingBaseModel}
D(q)\ddot{q} + H(q,\dot{q}) = Bu + J_{sp}(q)^\top \tau_{\rm sp}+ J(q)^\top \lambda,
\end{equation}
with $q$ the vector of generalized coordinates given in \eqref{eq:qdef}, $u$ the vector of motor torques, $\tau_{\rm sp}$ the spring torques, and $\lambda$ is the contact wrench. $D(q)$ is the mass-inertia matrix, $H(q,\dot{q})$ contains the velocity and gravitational terms, and $B$ is the motor torque distribution matrix. The Jacobian for the springs is given by
\begin{equation}
\label{eq:SpringJacobian}
J_{sp}(q):= \frac{\partial}{\partial q}\left[ \begin{array}{l} q_{ \rm 5L} \\q_{\rm 4L} + q_{\rm 5L} + q_{\rm 6L}\\  q_{ \rm 5R} \\q_{\rm 4R} + q_{\rm 5R} + q_{\rm 6R}\end{array} \right].
\end{equation}
The ground contact Jacobian is specified in the next subsection.

The open-source package FROST \cite{hereid2017frost} is used to generate all of the terms in \eqref{eq:FloatingBaseModel}. The floating base model used for control design is available as open-source code \cite{cassiemodel}. In that model, the   spring stiffness is taken to infinity, resulting in \eqref{eq:SpringRelaxed} holding. For simulations, the springs are kept.

\subsection{Hybrid Model for Walking}

Walking is modeled as a hybrid system corresponding to alternating phases of single support (one leg in contact with the ground and no slipping) and double support (both legs in contact with the ground). The double support phase is assumed to be instantaneous. This paper uses two domains; one for the right leg in stance, and one for the left leg in stance. The non-stance foot is also called the swing foot. Periodic gaits are computed over one step each on the right and left legs. The following assumes the \textit{right leg is stance} and the \textit{left leg is swing}. The labels $\rm R$ and $\rm L$ are swapped when the left leg is in stance.

\subsubsection{Single Support}

Ground contact is assumed to give rise to a holonomic constraint on foot position and orientation, $c_{\rm R}(q)$, that maintains the Cartesian position as well as the yaw and pitch angle of the stance foot to constant values during a step, while the foot's roll angle is unconstrained due to the narrow width of Cassie's feet. 
\begin{equation} \label{eq:holonomicConstraintsCR}
c_{\rm R}(q)^\top:= \left[p^{x}_{\rm R}, p^{y}_{\rm R}, p^{z}_{\rm R}, \psi^{\rm yaw}_R , \psi^{\rm pitch}_R \right]^\top.
\end{equation}
With the right foot in stance, the contact Jacobian is $J_{\rm R}(q):=\frac{\partial c_{\rm R}(q)}{\partial q}$; the contact constraint is imposed by setting its acceleration to zero
\begin{equation} \label{eq:holonomicConstraints}
J_{\rm R}(q)\ddot{q} + \dot{J}_{\rm R}(q,\dot{q})\dot{q} = 0.
\end{equation}
The contact Jacobian has full row rank of five, so using \eqref{eq:holonomicConstraints} it is possible to eliminate five coordinates from \eqref{eq:FloatingBaseModel} and the ground contact forces on the stance foot (the forces on the swing foot are identically zero), thereby obtaining a Lagrangian model with 15 DoF. We skip this step because later, for gait design, we once again use the package FROST which works directly with differential algebraic equations, such as \eqref{eq:FloatingBaseModel} and \eqref{eq:holonomicConstraints}.

\subsubsection{Double Support}
\begin{figure}[t]
\centering
     \subfloat[\label{fig:constraint}]{
     	\includegraphics[width=0.4\columnwidth]{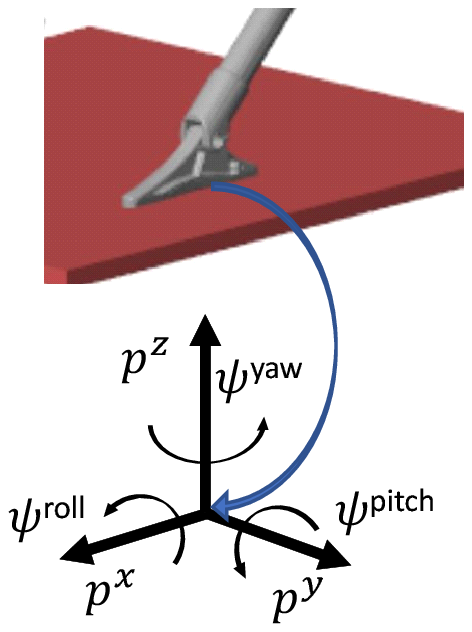}
	}~~~
    \subfloat[\label{fig:pitchlength}]{
      	\includegraphics[width=0.4\columnwidth]{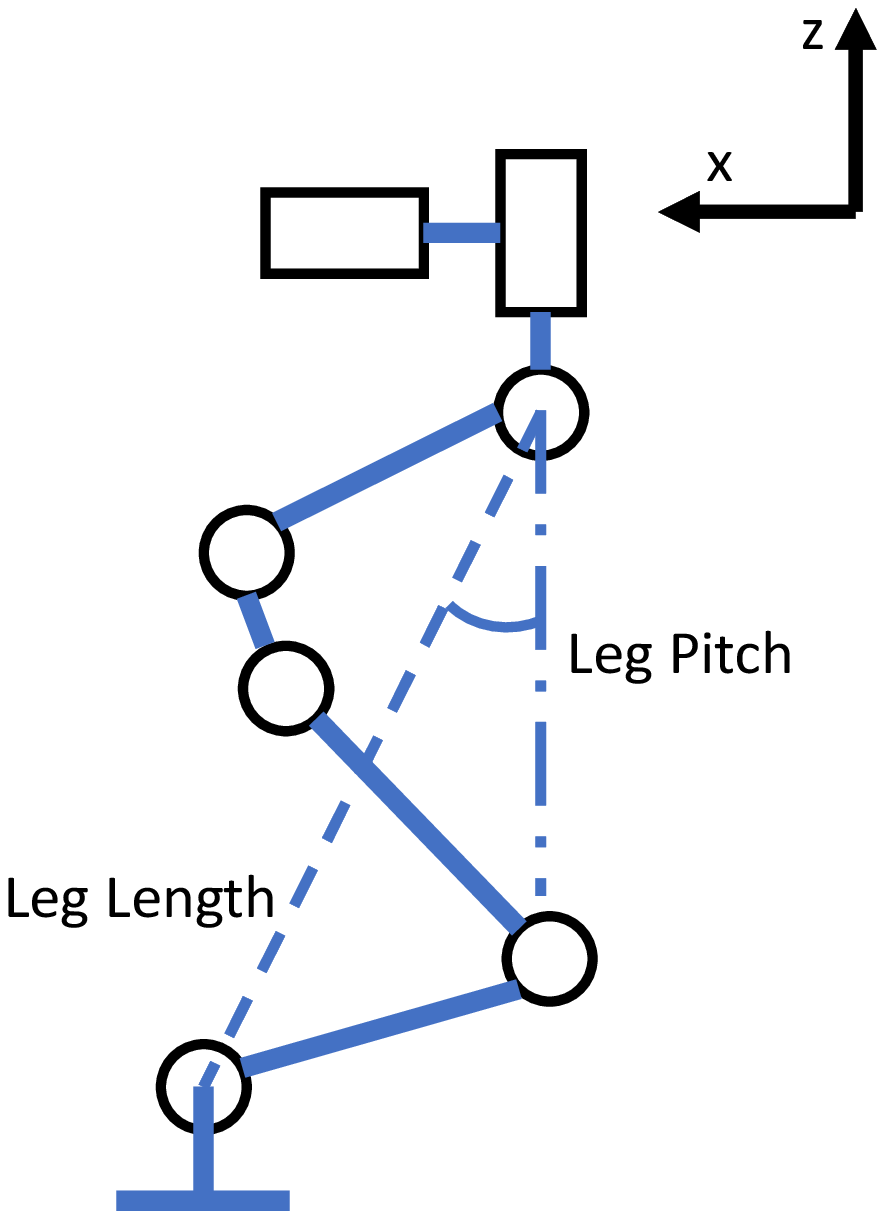}
      	}
     \caption{(a) The coordinates on the feet are shown. When a foot is in contact with the ground, there are five independent constraints leaving only the roll angle of the foot free. (b) The virtual leg is the dotted line from the hip to the top of the foot; its length is called Leg Length. The relative leg pitch is its angle of the virtual leg relative to the hip while absolute leg pitch is the relative angle plus the pitch angle of the  torso.}     
     \squeezeup \squeezeup
\end{figure}

The transition from single support to double support is captured by the height of the swing foot from the ground decreasing to zero. Specifically, with the right foot in stance, the transition set is
\begin{equation} \label{eq:guard}
{\cal S}_{\rm R \to L} = \{ (q,\dot{q}) \in \mathcal{TQ} \;|\; p^{z}_{L}(q) = 0,  \; \dot{p}^{z}_{L}(q,\dot{q}) < 0 \}
\end{equation}
where $p^z_{L}(q)$ is the vertical component of the position of the swing foot; see Fig.~\ref{fig:constraint}. 

\remark{During a step, the controller will maintain the pitch angle of the swing foot approximately level with the ground. Moreover, the applied torque is limited to a sufficiently low value that on solid ground, the weight of the robot easily and quickly places the foot level with the ground. Finally, impact is detected by the compression of the springs in joints $q_{5L}$ and $q_{6L}$.}
 
The instantaneous impacts are modeled through a discrete map that results in a discontinuity in the velocity of the system $\dot{q}^-$ just before impact and the velocity of the system $\dot{q}^+$ just after impact, while the positions do not change \cite{westervelt2007feedback}. 
Moreover, just after impact, the former swing foot is assumed to satisfy the same constraints as those imposed on the stance foot. Letting $c_{\rm L}(q)$ denote the corresponding holonomic constraint, the pre- and post-impact velocities then satisfy
\begin{equation} \label{eq:impactMap}
\begin{bmatrix}
D(q)  &  -J_{\rm L}^T(q) \\
J_{\rm L}(q)  & 0 \\
\end{bmatrix}
\begin{bmatrix}
\dot{q}^+ \\
\delta F_{\rm L} \\
\end{bmatrix}
=
\begin{bmatrix}
D(q)\dot{q}^- \\
0 \\
\end{bmatrix},
\end{equation}
where $\delta F_{\rm L}$ is the vector of contact impulses. Because the Jacobian $J_{\rm L}(q)$ has full row rank and $D(q)$ is positive definite, the left hand side of \eqref{eq:impactMap} is invertible. Projecting the solution of \eqref{eq:impactMap} to the velocity components defines the impact map, 
\begin{equation} 
\label{eq:Delta}
\dot{q}^+=:\Delta_{\rm R \to L}(\dot{q}^-).
\end{equation}

\subsubsection{Hybrid Model}
The overall model is given as follows:
\begin{equation}
\label{eq:model:hcs}
\Sigma : \left\{
    \begin{aligned}\smallskip 
   D(q)\ddot{q} + H(q,\dot{q}) &= Bu + J_{sp}(q)^\top \tau_{\rm sp}+ J_{\rm R}(q)^\top \lambda   \\
   J_{\rm R}(q)\ddot{q} + \dot{J}_{\rm R}(q,\dot{q})\dot{q} &= 0 \hspace*{1.6cm} (q;\dot{q}^-) \not \in {\cal S_{\rm R \to L}} \\
   \dot{q}^+&=\Delta_{\rm R \to L}(\dot{q}^-)~~ (q;\dot{q}^-) \in {\cal S_{\rm R \to L}} \\ 
   \\
    D(q)\ddot{q} + H(q,\dot{q}) &= Bu + J_{sp}(q)^\top \tau_{\rm sp}+ J_{\rm L}(q)^\top  \lambda  \\
   J_{\rm L}(q)\ddot{q} + \dot{J}_{\rm L}(q,\dot{q})\dot{q} &= 0 \hspace*{1.6cm} (q;\dot{q}^-) \not \in {\cal S_{\rm L \to R}} \\
   \dot{q}^+&=\Delta_{\rm L \to R}(\dot{q}^-) ~~ (q;\dot{q}^-) \in {\cal S_{\rm L \to R}}.    
      \end{aligned}
      \right.
\end{equation}

\subsection{Model for Standing}
Cassie can stand quietly. The model is given by \eqref{eq:FloatingBaseModel} with $J^\top=[J_{\rm R}^\top, J_{\rm L}^\top]$. If the feet are flat on the ground, 10 DoF are removed. In this case, when the spring stiffness is considered to be infinite, and the $x$-axes of the two feet do not lie on a common line, the model is fully actuated.


%% file: Sections/WalkingController.tex
\section{Walking on Various Terrains}
\label{sec:Walking}

This section presents the initial walking gait controller implemented on Cassie. The control design is based on the method of \textit{virtual constraints}, a \textit{gait library} limited to the sagittal plane, and \textit{leg-angle adjustment} in the sagittal and frontal planes \cite{da20162d}. 

\subsection{Virtual Constraints}

Virtual constraints are functional relations among the generalized coordinates of the robot that are asymptotically imposed on the system through feedback control. In particular, the virtual constraints are expressed as outputs of the model \eqref{eq:FloatingBaseModel} in the form
\begin{equation} 
\label{eq:VirtualConstraint} 
y = h(q, \tau, \alpha) = h_0(q) - h_d (\tau,\alpha),
\end{equation}
where $h_0$ specifies the quantities being regulated, $h_d$ encodes their desired evolution, and $\alpha$ is a matrix of real coefficients that parameterizes the spline $h_d$. The phasing variable, $\tau$, satisfies
\begin{align}
\dot{\tau} &= \frac{1}{T}, & (q;\dot{q}^-) \not \in  {\cal S_{\rm R \to L}} \cup  {\cal S_{\rm L \to R}} \nonumber\\
\tau^+&=0 & \text{otherwise},
\end{align} 
where $T$ is the nominal step duration.  

A controller is then designed with the objective of zeroing the outputs, i.e., $y\equiv 0$, thereby achieving the virtual constraints. The zeroing of the output value will be at best accomplished asymptotically, and in practice, on a physical robot, only approximately. 


\subsection{Choice of What to Control}

The most direct choice for the regulated quantities, $h_0$, would be the actuated joints of Cassie, which are a subset of the body coordinates. On previous planar robots we have studied, such as Rabbit and MABEL \cite{WEBUGR04,park2013finite}, this made sense because these robots had a simple (human-inspired) morphology and the control objectives could be  associated in an intuitive manner with hip and knee angles. On our first 3D biped, MARLO, the legs, and hence the actuated coordinates in the sagittal plane, were associated with four-bar linkages, which gave rise to synthesizing coordinates that were associated with a virtual leg connecting the hip to the end of the leg \cite{GriffinIJRR2016}. MARLO also had a rather tall torso that provided adequate inertia about the roll axis so that adjustments by the stance leg hip motor to maintain the torso roll angle approximately zero would not cause oscillations on the swing leg roll angle. 

For the land-bird-inspired Cassie, if one is not a biologist, the actuated joints have limited physical meaning with respect to the walking behavior of the robot. Hence, we choose instead to regulate torso orientation, stance and swing leg lengths, swing leg orientation, and swing foot pitch angle. We find these quantities to be rather universal across  bipedal platforms (whether human or bird inspired) and directly relatable to gait outcomes. For example, leg lengths are directly related to the height of the torso and foot clearance; swing leg pitch and roll angles at impact are commonly used in bipeds to  regulate walking speed, and the yaw angle of the swing leg sets the direction of the robot for the next step.  Moreover, because Cassie is roughly a basketball-sized sphere with one meter-long legs attached to it, its (spherical) torso provides little ``mechanical filtering'' between the motion of the stance and swing legs. 


With this in mind, we define the following nine outputs, with the stance foot pitch angle left passive\footnote{For now, it is simply observed that leaving it passive avoids ``foot roll''; when walking on soft sand, we will come back to this feature.},
\begin{equation}
\label{eq:h0}
h_0(q) = \left[ \begin{array}{l}  q_{\rm roll} \\ q_{\rm 2~st} \\q_{ \rm pitch} \\ q_{\rm LL~st} \\ q_{\rm LR~sw} \\ q_{\rm 2~sw} \\ q_{\rm LP~sw} \\ q_{\rm LL~sw} \\ q_{\rm FP~sw} \end{array} \right]
~~~\left( \begin{array}{c}  \text{torso roll} \\ \text{stance hip yaw} \\ \text{torso pitch} \\ \text{stance leg length} \\ \text{swing leg roll}  \\ \text{swing hip yaw}  \\ \text{swing leg pitch}  \\ \text{swing leg length}  \\ \text{swing foot pitch}  \end{array} \right),
\end{equation}
where, when the right leg is stance and the left leg is swing, 
\begin{align}
\label{eq:AbsolteQuantities}
  q_{\rm LL~st} =& \sqrt{0.5292 \cos(q_{\rm 4R}+0.035)+0.5301}  \nonumber\\
q_{\rm LR~sw} =& q_{\rm roll} + q_{\rm 1L}  \nonumber \\
  q_{\rm LP~sw} =& -q_{\rm pitch} + q_{\rm 3L}  \\
  - \arccos( & \frac{0.5(\cos(q_{\rm 4L} + 0.035) + 0.5292)}{\sqrt{0.5292 \cos(q_{\rm 4L}+0.035)+0.5301}})+0.1 \nonumber \\ 
  q_{\rm LL~sw} =& \sqrt{0.5292 \cos(q_{\rm 4L}+0.035)+0.5301}  \nonumber\\
      q_{\rm FP~sw} =& -q_{\rm pitch} + q_{\rm 7L} + 1.1. \nonumber
\end{align}
The forward kinematics of leg length and leg pitch are calculated based on configuration of the robot and shown in Fig. \ref{fig:pitchlength}. For clarity, leg pitch refers to the absolute pitch angle of the virtual leg when torso roll and yaw are zero; leg roll is defined in a similar manner.

For later use in control implementation, we note that the to-be-regulated quantities in \eqref{eq:AbsolteQuantities} can be expressed in terms of the \textit{actuated joints} via
\begin{align}
\label{eq:IK}
q_1 =& q_{\rm LR} - q_{\rm roll}   \nonumber \\
q_3 =&    q_{\rm LP} + q_{\rm pitch}  \nonumber \\
& +\arccos \left( \frac{0.9448 ~q_{\rm LL}^2 -0.0284}{q_{\rm LL}}\right)-0.1 \nonumber \\
q_4 =& \arccos\left(1.8896~q_{\rm LL}^2 - 1.0017\right) - 0.035 \\
    q_7 =& q_{\rm FP} + q_{\rm pitch} - 1.1 \nonumber,
\end{align}
where the distinction between stance and swing has been dropped.


\subsection{Set of Gaits for Walking at Various Speeds}
The desired evolution of the virtual constraints is defined by $h_d$ in the output equation \eqref{eq:VirtualConstraint}. This function is constructed using linear interpolation of a discrete library of gaits, each encoding a particular forward walking speed. Here, seven gaits were generated where the average velocity in the sagittal plane, $\bar{v}_x$, ranged from -0.5~m/s to +1.0~m/s in 0.25~m/s increments. We assume that the virtual constraints of  each of these ``open-loop'' gaits have a desired trajectory, $h_d^i(\tau, \alpha_i)$, that is parameterized by a set of 5\textsuperscript{th}-order B\'{e}zier polynomials with the corresponding matrix of coefficients denoted as $\alpha_i$. The step time for all gaits was chosen to be a constant. Trajectory optimization is then used to independently solve for each $\alpha_i$.

The nonlinear optimization problems were constructed and solved using FROST \cite{hereid2017frost}, which internally uses the direct-collocation trajectory optimization framework developed by Hereid et. al. \cite{hereid20163d}. Each hybrid optimization was performed over two domains (right stance then left stance), where the following cost function was minimized:
\begin{equation*}
\begin{split}
\text{Domain Cost} &= \int_{\tau=0}^{\tau=1} \big(||u||^2 + c\,|q_{pitch}|^2 + c\,|q_{roll}|^2 \\
                               &+ c\,|q_{\rm 1L}|^2 + c\,|q_{\rm 2L}|^2 + c\,|q_{\rm 1R}|^2 + c\,|q_{\rm 2R}|^2\big) \; d\tau.
\end{split}
\end{equation*}
The addition of the torso pitch/roll and the hip roll/yaw angles into the cost function (multiplied by a large weight, $c=10,000$) guides the optimizer to find gaits with minimal roll and yaw movement. Constraints are placed on the optimization problem to ensure that the optimized gait is periodic over two steps and that the left and right stance are symmetric\footnote{Due to the enforced symmetry, it is possible to write this optimization problem using a single domain. However, the general two-step formulation allows for future design of non-symmetric gaits through simple removal of this particular constraint.}. Torque, joint angle, and joint velocity limits were imposed to ensure that the gait can be physically-realized on the actual robot. Additional constraints are outlined in Table \ref{table:optimization_constraints}.

\begin{table}[h!]
\centering
\begin{tabular}{|l l|} 
 \hline
 Average sagittal velocity, $\bar{v}_x$   &= $v_i$~m/s \\
 Average lateral velocity,  $\bar{v}_y$   &= 0~m/s \\
 Step time                                   &= 0.4~s \\
 Torque for stance foot pitch                &= 0~Nm \\
 Friction cone, $\mu$                        &< 0.6 \\
 Mid-step swing foot clearance               &> 0.15~m \\
 Absolute swing foot pitch                   &= 0~rad \\
 Distance between feet                       &> 0.2~m \\
 Distance between pelvis and stance foot     &$\in$ (0.5, 1)~m \\
 Swing foot velocity on impact ($x$ and $y$) &= 0~m/s \\   
 Swing foot velocity on impact ($z$)         &$\in$ (-1, 0)~m/s \\     
 \hline
\end{tabular}
\caption{Constraints used in gait optimizations. For each of the seven optimizations, the average sagittal velocity was constrained to a different value between $-0.5$ and $1$ m/s.}
\label{table:optimization_constraints}
\end{table}


Each of the 7 optimization problems yields a single parameter matrix, $\alpha_i$, and takes approximately 3~min to solve using IPOPT in MATLAB. 

\remark{A C++ implementation of the optimization problem formed by FROST has been posted on GitHub \cite{hereid2018rapid}. It allows parallel computation of the gaits and cloud-based gait optimization.}

\subsection{Approximately Implementing the Virtual Constraints}
\label{sec:PDControl}


If the overall dynamic model and joint angular velocity estimates were sufficiently accurate, we could implement the virtual constraints via input-output linearization. Indeed, the outputs \eqref{eq:VirtualConstraint} have relative degree two \cite{ISIDORIA95} and the row rank of the decoupling matrix is full rank on the control-design model. 

On the actual robot, however, the power amplifiers, motor dynamics, network delays, and walking surface are not adequately characterized to allow model-based torque control (the mechanical model itself is not the main source of uncertainty). Consequently, the virtual constraints are approximately imposed through decoupled PD controllers, as in \cite{WEBUGR04,park2013finite,GriffinIJRR2016}. To do this, \eqref{eq:IK} is used to rewrite \eqref{eq:VirtualConstraint}  as
\begin{equation}
\label{eq:Constraints4Implementation}
\widetilde{y}=\widetilde{h}_0(q)-\widetilde{h}_d(\tau, q_{\rm pitch}, q_{\rm roll}, \alpha),
\end{equation}
with (right leg in stance)
\begin{equation}
\label{eq:h0tilde}
\begin{aligned}
\widetilde{h}_0(q)^\top =& \left[  q_{\rm roll}, q_{\rm 2R}, q_{ \rm pitch}, q_{\rm 4R}, \right.\\
&\left. ~~q_{\rm 1L}, q_{\rm 2L}, q_{\rm 3L}, q_{\rm 4L}, q_{\rm 7L} \right]^\top
\end{aligned}
\end{equation}
and
\begin{equation}
\label{eq:hdtilde}
\widetilde{h}_d(\cdot) = \left[ \begin{array}{l} h_{d~1}(\cdot) \\
h_{d~2}(\cdot) \\
h_{d~3}(\cdot) \medskip \\
\arccos\left(1.8896~\left[h_{d~4}(\cdot)  \right]^2 - 1.0017\right) - 0.035 \medskip \\
h_{d~5}(\cdot) - q_{\rm roll} \\
h_{d~6}(\cdot) \\
 h_{d~7}(\cdot) + q_{\rm pitch}  +0.1 \medskip \\
 ~~+\arccos \left( \frac{0.9448 ~\left(h_{d~8}(\cdot)\right)^2 -0.0284}{h_{d~8}(\cdot)}\right)\medskip \\
\arccos\left(1.8896~\left[h_{d~8}(\cdot)\right]^2 - 1.0017\right) - 0.035 \medskip \\
h_{d~9}(\cdot) + q_{\rm pitch} - 1.1 
 \end{array} \right].
\end{equation}
\remark{Though not proven here, \eqref{eq:Constraints4Implementation} implements the same virtual constraints as \eqref{eq:VirtualConstraint} and
\eqref{eq:AbsolteQuantities}; one is zero if, and only if, the other is zero.}

\begin{figure}[t]
\vspace{.25cm}
	\centering
	\includegraphics[width=1\columnwidth]{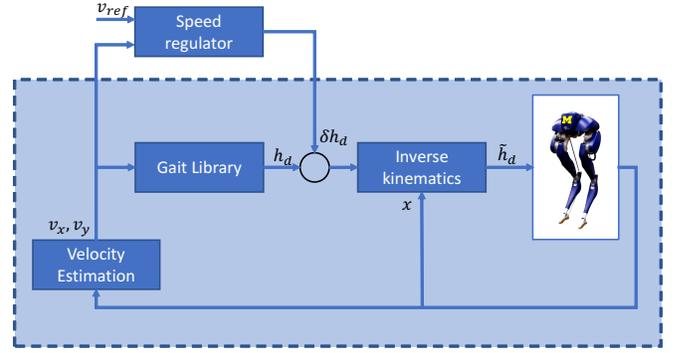}
	\caption{Control Diagram for Walking. The feedback loop implementing the virtual constraints and the gait library (blue box) maintains the robot's posture and synchronizes the legs for walking. Moreover, the resulting closed-loop system renders the dynamics of the center of mass velocity close enough to that of an inverted pendulum that it can be regulated by adjusting the pitch and roll angles of the swing leg.}
	\label{fig:control_diagram}
\squeezeup \squeezeup
\end{figure}

In \eqref{eq:h0tilde} and \eqref{eq:hdtilde}, the outputs are ordered so that they correspond to the first four actuators on the stance leg followed by the five actuators on the swing leg. Recall that the torque on the stance foot is set to zero. The virtual constraints are approximately zeroed with a classical PD controller,
\begin{equation}
\label{eq:ytildecontrol}
u= -K_{P} \widetilde{y} - K_{D}\dot{\widetilde{y}}, \\
\end{equation}
where the $9 \times 9$ matrices $K_{P}$ and $K_{D}$ are diagonal.\\
\remark{The torso pitch and roll angles are world frame coordinates. When the stance foot is firmly on the ground (i.e., not slipping), they can be controlled through the hip motors. The remaining outputs are directly actuated.  When the springs are ignored, the output has vector relative degree two.}

\subsection{Gait Library and Stabilization by Leg Angle Adjustment}

The \textit{Gait Library} is an interpolation of the seven discrete gaits into a continuum of gaits valid for $-0.5 \le v_{x} \le 1.0$ m/s. The interpolation parameter is the robot's filtered sagittal velocity. The implementation of the gait library is done exactly as in \cite[Eq.~(8)-(10)]{da20162d} and does not introduce any new parameters into the controller. 

With the gait library implemented, the closed-loop system is unstable in the sense that the $(\dot{x},\dot{y})$ Cartesian coordinates are approximate integrators; see \cite[Sect.~III-C]{da20162d} for the explanation. Leg angle adjustment is added to stabilize the closed-loop system. The implementations for longitudinal and lateral  velocity stabilization are based on \cite[Eq.~(13) and (17)]{da20162d}. These controllers add four more control parameters. The overall control strategy is shown in Fig.~\ref{fig:control_diagram}.


\subsection{Parameter Tuning}
\label{sec:ParamTuning}

The right and left legs of the robot are sufficiently symmetric that control parameters for the left and right legs are the same. 
The controller was implemented in \RTSimulink and the parameters tuned by hand on a \SimMech model provided to us by Agility Robotics. A process for tuning the 18 joint-level PD parameters and the 4 leg-angle PD parameters on the robot is posted with the code on GitHub. 

\subsection{Experiments}
\label{sec:walkingExperiments}

A first version of the walking controller was implemented on Cassie Blue six weeks after arrival on campus and was demonstrated to the Associated Press (AP) on October 23rd, 2017 \cite{CassieOnAP}. On June 2, 2018, we damaged a leg on Cassie Blue and sent her back for repairs. While the robot was in the shop, Agility Robotics upgraded the hip roll and yaw joints to match those on their current production model, significantly reducing friction in them. At the same time, Agility also modified the MATLAB environment in which a user's controller is implemented, breaking our controller. Because we would be soon modifying the robot with the addition to a 15 kg torso, we did not spend much time retuning the controller.

The remainder of the section discusses some of the many terrains on which the robot has been challenged to operate over the past 11 months as documented in \cite{CassieACC2019}. The dates of the experiments are noted below. In each experiment, the robot is being directed by an operator via an RC Radio with commands ``stand quietly'' or ``walk''. When walking, the robot is sent desired $\bar{v}_x^{tg}$, $\bar{v}_y^{tg}$, and turn rate. 

\subsubsection{Initial Testing in the Laboratory}

\begin{figure}[b]
\squeezeup \squeezeup
	\centering
	\includegraphics[width=0.7\columnwidth]{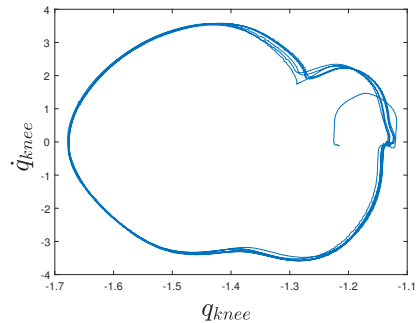}
	\caption{The phase portrait of the left knee when Cassie is walking in place, starting from a standing position. The units are $\rm rad$ and $\rm rad/s$.}
	\label{fig:phasePortrait}
\end{figure}

After the controller was successfully working in closed-loop with the \SimMech~ model, it was transferred to the robot and the PD gains tuned over a period of a few days. During this process, an overhead gantry was used to catch the robot in case of a fall. The gains were initially tuned for walking in place. Once that milestone was achieved, walking at various speeds came quickly. A typical limit cycle is shown in Fig.~\ref{fig:phasePortrait}.

\subsubsection{North Campus Grove}
Cassie Blue was taken outdoors for the first time and demonstrated to the Associated Press \cite{CassieOnAP} on 23 October 2017. Initially, a safety gantry was used. After walking on a sidewalk with no difficulty, Cassie was released from the gantry. After initial walking on a level concrete area, Cassie was steered onto the grass. Due to a nervous operator hitting the wrong button on the RC-controller, the robot sped up and had to be stabilized by a researcher. The softer nature of the grass caused no difficulties in detecting foot impacts via spring deflection. For the next 11 minutes, the robot was steered on and off grass and hard surfaces, onto sloped grass surfaces, turned, and walked in place. When the (unplanned) trajectory brought the robot and researchers near a grassy knoll, the decision was made to see if the robot could handle it. The robot unexpectedly sped up when heading up the slope, leaped over a bench, and fell at least 1.5~m onto the concrete. This put a slight dent in the battery, but caused no other observable damage as the robot was quickly rebooted and walked in place. This ended the experiment.

\subsubsection{Waxed Floors and Snow} Cassie Blue was taken to the UM Dental School on 11 December 2017 at the invitation of Dean Laurie McCauley. The video can be found here \cite{CassieDental}. The robot handled well on surfaces with reduced friction. An unplanned bump into the pillar and walking in the snow are shown here \cite{CassieACC2019}.

\subsubsection{Controlled Burn for Wild Grasses} On 22 April 2018, a controlled burn was conducted on the UM campus to promote the growth of native grasses. After clearing it with the personnel conducting the burn, Cassie Blue walked over sloped ground, in heavy smoke, and over short burning grass, branches, and leaves \cite{CassieFire}. The robot never fell. While the exercise demonstrated our general confidence in the robot's controller, it was done to drive home the fact that a battery-powered robot does not suffer from smoke inhalation and can take some heat. We'll return in 2019 with a full perception package on the robot.

\subsubsection{Sand Volley Ball Court} On 09 May 2018, the Discovery Channel filmed Cassie Blue. The Segway riding, reported later, was their main interest. Since we were near the sand volley ball court, we challenged the robot to walk on it \cite{CassieSand}. The narrow feet sunk into the sand, with the ``heel'' digging in the most. Because the stance foot is passive, the robot's gait remained quite stable. The robot walked more slowly than on grass (possibly due to foot slip) and it traversed the entire course, passing under the net. For the second pass, a pair of tennis shoes was placed on the robot. Cassie then walked no differently than when on grass or concrete.

\begin{figure}
     \subfloat[Grass\label{subfig:grass}]{
     	\includegraphics[height=3.3cm]{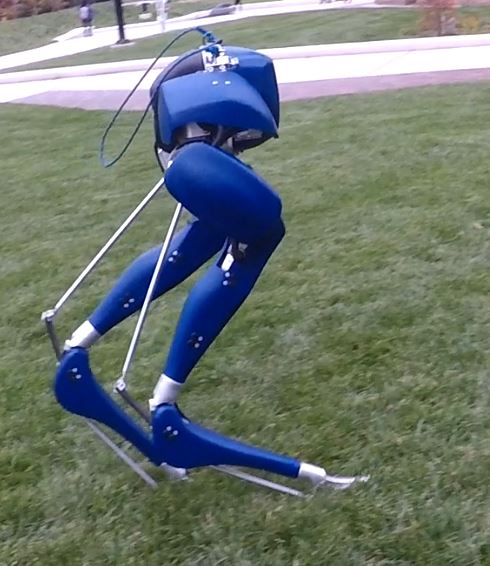}
        }
      \subfloat[Burning Brush\label{subfig:fire}]{
      	\includegraphics[height=3.3cm]{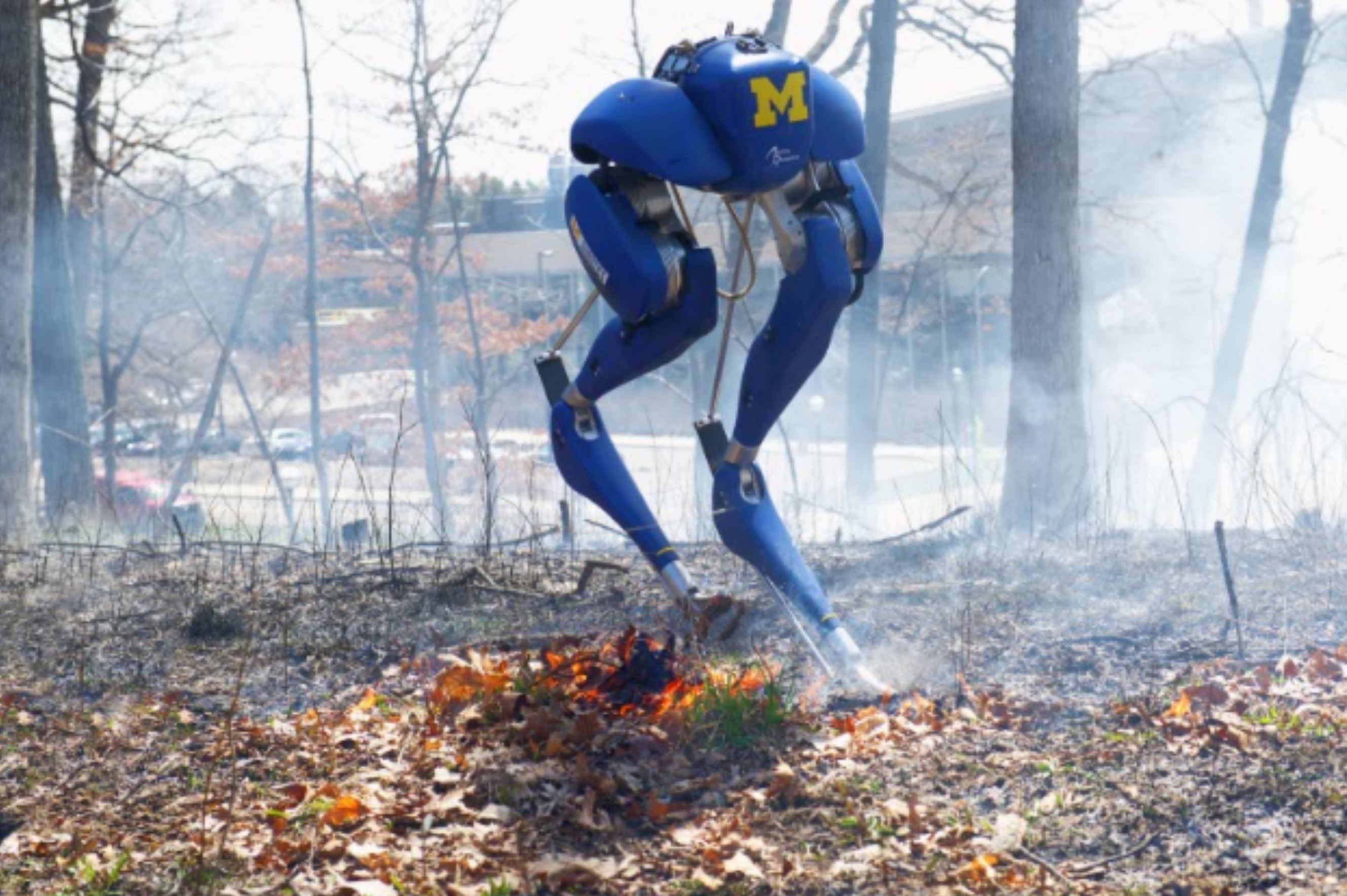}
      	}\\
      \subfloat[Snow\label{subfig:snow}]{
     	\includegraphics[height=3.7cm]{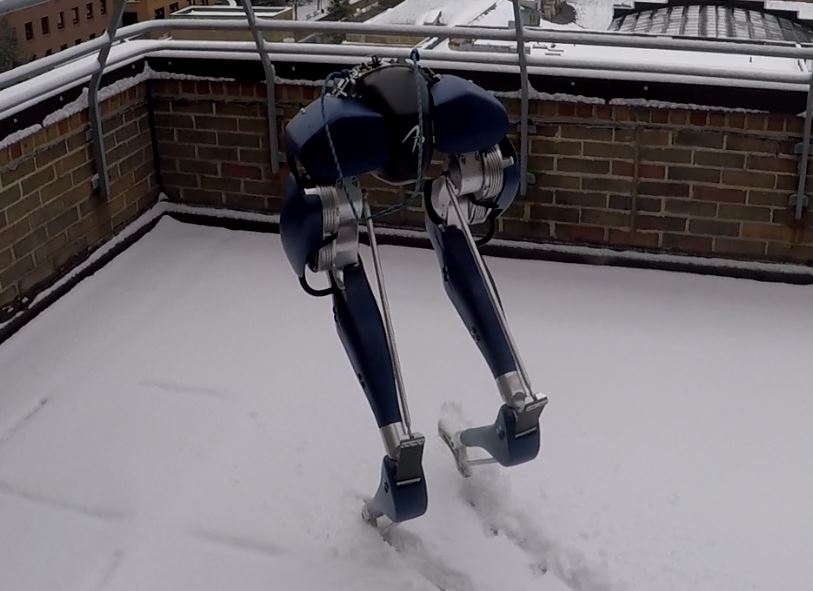}
        } 
     \subfloat[Sand\label{subfig:sand}]{
     	\includegraphics[height=3.7cm]{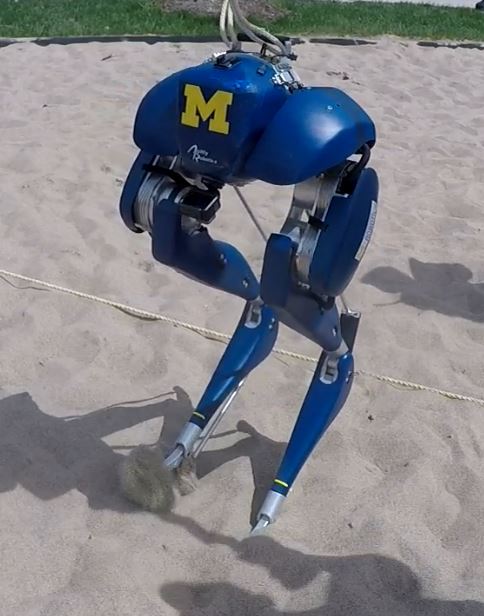}
        }
        \caption{Cassie Blue walking on various unmodeled terrains.}
           \squeezeup
      \vspace*{-3mm}
\end{figure}

%% file: Sections/Standing.tex
\section{Quiet Standing and Riding a Segway}
\label{sec:StandingAndSegway}

For standing, all ten actuators are used. Foot actuation is required to prevent rotation of the robot about the $y$-axis in the body frame. The standing controller also allows Cassie to ride a Segway; yes, a robot riding another robot. 

\subsection{Quiet Standing}
\label{sec:Standing}

The standing condition is here assumed to be reached either from a stepping-in-place gait or by a user booting up the robot with the torso suspended approximately a half meter off the ground. The feet are assumed to be flat on the ground and beneath the torso. 

Let $p^{\rm CoM} = (p^{\rm CoM}_x, p^{\rm CoM}_y, p^{\rm CoM}_z)$ be the center of mass of the robot. With the feet flat on the ground and $(p^{\rm CoM}_x, p^{\rm CoM}_y)$ within the convex hull of the feet, the robot can maintain a static pose. To maintain the desired center of mass position, $p^{\rm CoM}_x$ is regulated with the actuators for the pitch angle of the feet, while $p^{\rm CoM}_y$ is regulated in an indirect way: the roll angle, instead of $p^{\rm CoM}_y$, is controlled to be zero. The roll is controlled by adjusting the leg length difference in the two legs. A zero roll angle is equivalent to a centered $p^{\rm CoM}_y$ when all other joints are symmetric. One reason we are using roll angle for feedback is because its value is less noisy than $p^{\rm CoM}_y$, which is calculated via a kinematic chain. Another reason is that if the ground is sloped in the frontal plane, causing the robot to lean to the right\footnote{This feature is useful even on flat ground. The PD control of leg length should be thought of as soft springs. A bit of lean to the right places more weight on the right leg, which causes further compression of the spring in the right leg, which causes further leaning, etc., until $p^{\rm CoM}_y$  moves to the right of the foot and the robot falls. With high PD gains, this cascading effect can be avoided, but regulating $p^{\rm CoM}_y$ solves the problem with lower gains.}, the right leg will automatically be extended and the left leg compressed, as shown in Fig.~\ref{fig:CassieSegway}. The height of the standing pose $p^{\rm CoM}_z$ is set by adjusting the average of the two leg lengths. The hip roll and yaw joints are regulated to constant values. With this controller, and the two feet roughly 0.3~m apart, Cassie is able to squat almost flat on the ground and stand to a height of approximately one meter. Quiet standing, lowering to a squat, and standing back up are illustrated in the video associated with the paper \cite{CassieACC2019}.
\subsection{Riding a Segway}
\label{sec:Segway}
\begin{figure}
	 \centering
     \subfloat[]{
     	\includegraphics[height=4cm]{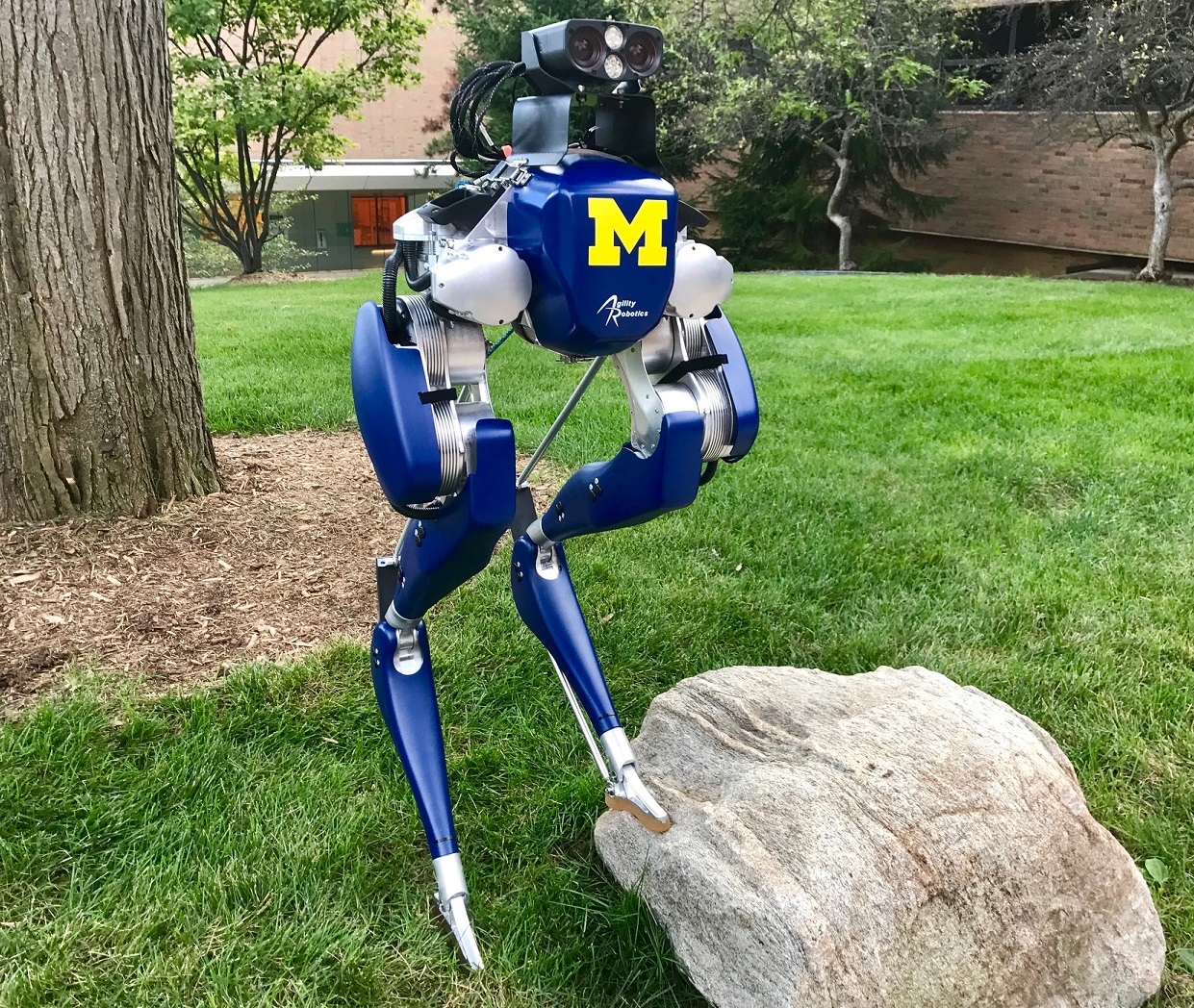}
        }
      \subfloat[]{
      	\includegraphics[height=4cm]{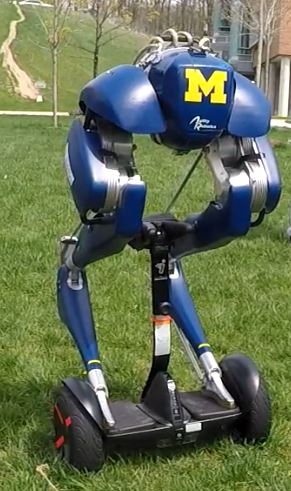}
      	}
	  \caption{(a) Cassie standing on an uneven surface. (b) A stack of robots. Cassie is riding a Segway miniPro. The Segway will accelerate forward if the foot platform leans forward. Turning is controlled by pushing against the central vertical bar.} 
      \squeezeup
      \vspace*{-3mm}
	  \label{fig:CassieSegway}
\end{figure}

Figure~\ref{fig:CassieSegway} shows Cassie Blue riding a Segway miniPro (robot).  
The dynamics of the Segway are unknown and its states are not measured. The acceleration and direction of the Segway are determined by body lean, that is, by adjusting the target center of mass position. As elsewhere, the commands are sent by an operator via radio control.

To accelerate or decelerate, $p^{\rm CoM}_x$ is shifted forward or backward, respectively. To turn, Cassie needs to lean into the center bar with her legs, which is accomplished by shifting $p^{\rm CoM}_y$. With the nominal standing controller, the Segway would oscillate when Cassie was placed on its platform. The feedback gains on the feet were turned down and the oscillations ceased.  

With the crouched posture seen in Fig.~\ref{fig:CassieSegway}, Cassie Blue was able to ride on sidewalks and grassed areas at roughly 4 m/s \cite{CassieSegway}.  To be clear, the robot was placed on the Segway by an operator. Mounting and dismounting the Segway were not addressed. 

%% file: Sections/Conclusion.tex
\vspace{-1mm}
\section{Conclusion}
\label{sec:conclusion}



The Cassie biped has the morphology of a large flightless bird.  The full 20 DoF dynamic model and optimization were used to design seven gaits for walking in place, forward, and backward, while meeting key physical constraints. The complicated morphology of the robot was translated into ``universal,'' physically meaningful control objectives involving torso orientation, leg orientation, and leg length. Moreover, it was shown how to practically implement these control objectives via decoupled PD controllers on the robot. 

The gait-library method of \cite{da20162d} was used to gain-schedule the discrete gaits into a continuum of walking gaits. The final controller was demonstrated both in and out of the laboratory, including walking on sidewalks, grass, sand, waxed floors, snow, and a hill with short brush \cite{CassieACC2019}. The control software will be made open-source if the paper is accepted for publication. 


In terms of next steps, we are working on a data-driven method to design feed forward torques to improve tracking performance. We are also seeking gaits that allow fast motion with turning, perhaps using the method introduced in \cite{DaGr17IJRR}.

\balance